\pdfoutput=1

\documentclass[11pt]{article}

\usepackage[]{EMNLP2022}

\usepackage{times}
\usepackage{latexsym}

\usepackage[T1]{fontenc}

\usepackage[utf8]{inputenc}

\usepackage{microtype}

\usepackage{inconsolata}

\usepackage{algorithm}
\usepackage{algorithmic}
\usepackage{microtype}
\usepackage{amsmath}
\usepackage{multirow}
\usepackage{amsmath,array,graphicx,amssymb}
\usepackage[export]{adjustbox}
\usepackage{subcaption}
\usepackage{multirow}
\usepackage{hyperref}
\usepackage{mathtools} 
\usepackage{tabularx}
\usepackage{breakcites}
\usepackage{xcolor}
\usepackage{array}
\usepackage{makecell}
\usepackage{tabularx}
\usepackage{mathrsfs}
\usepackage{scalerel}
\usepackage{soul}
\usepackage{colortbl}
\definecolor{mygray}{gray}{.95}
\usepackage{soul}
\usepackage{booktabs}

\usepackage{graphicx}
\usepackage{caption}
\usepackage{algorithm}
\usepackage{algorithmic}
\usepackage{microtype}
\usepackage{amsmath}
\usepackage{multirow}
\usepackage{amsmath,array,graphicx,amssymb}
\usepackage{caption}
\usepackage{xspace}
\usepackage{amsthm}
\newtheorem{theorem}{Theorem}

\newtheorem{definition}{Definition}
\newtheorem{innercustomthm}{Theorem}
\newenvironment{customthm}[1]
  {\renewcommand\theinnercustomthm{#1}\innercustomthm}
  {\endinnercustomthm}

\newcommand{\tool}{\textsc{Jft}\xspace}

\usepackage{soul}
\usepackage{color}
\usepackage{xcolor}

\definecolor{apricot}{RGB}{255,240,234}
\definecolor{amber}{RGB}{214,81,30}

\definecolor{lightcyan}{RGB}{229,250,245}
\definecolor{teal}{RGB}{0,121,86}

\DeclareRobustCommand{\hlapricot}[1]{{\sethlcolor{apricot}\hl{#1}}}
\DeclareRobustCommand{\hllightcyan}[1]{{\sethlcolor{lightcyan}\hl{#1}}}
\usepackage{tikz}

\newcommand{\wyshi}[1]{\textcolor{black}{#1}}

\title{Just Fine-tune Twice: Selective Differential Privacy for Large Language Models}

\author{Weiyan Shi$^{\dagger}$, Ryan Shea$^{\dagger}$, Si Chen$^{\ddag}$, Chiyuan Zhang$^{\diamond}$, Ruoxi Jia$^{\ddag}$, Zhou Yu$^{\dagger}$\\
Columbia University$^{\dagger}$, Virginia Tech$^{\ddag}$, Google Research$^{\diamond}$ \\
\texttt{\{ws2634,rs4235\}@columbia.edu}, \texttt{chensi@vt.edu}, \texttt{chiyuan@google.com}, \\
\texttt{ruoxijia@vt.edu}, \texttt{zy2461@columbia.edu}}

\begin{document}
\maketitle
\begin{abstract}


Protecting large language models from privacy leakage is becoming increasingly crucial with their wide adoption in real-world products. Yet applying \emph{differential privacy} (DP), a canonical notion with provable privacy guarantees for machine learning models, to those models remains challenging due to the trade-off between model utility and privacy loss. Utilizing the fact that sensitive information in language data tends to be sparse, \citet{shi2021selective} formalized a DP notion extension called \emph{Selective Differential Privacy} (SDP) to protect only the sensitive tokens defined by a policy function. However, their algorithm only works for RNN-based models. In this paper, we develop a novel framework, \emph{Just Fine-tune Twice} (\textsc{Jft}), that achieves SDP for state-of-the-art large transformer-based models. Our method is easy to implement: it first fine-tunes the model with \emph{redacted} in-domain data, and then fine-tunes it again with the \emph{original} in-domain data using a private training mechanism. Furthermore, we study the scenario of imperfect implementation of policy functions that misses sensitive tokens and develop systematic methods to handle it. Experiments show that our method achieves strong utility compared to previous baselines. We also analyze the SDP privacy guarantee empirically with the canary insertion attack\footnote{Our code and data are available at  \url{https://github.com/wyshi/sdp_transformers}}.

\end{abstract}

\section{Introduction}



With the rapid advancement in natural language processing (NLP), 
it has become increasingly important to protect NLP models from leaking privacy information. 
Previous work has attempted to tackle this challenge by 
applying \emph{differential privacy}~\citep[DP,][]{dwork2014algorithmic} on these models 
\cite{mcmahan2018learning,li2021large}.  
However, existing DP learning algorithms suffer from \emph{limited user control} and \emph{low utility}, as they protect the entirety of each training example (e.g., one complete sentence) regardless of users' privacy preference, and therefore tend to be overly pessimistic when only partial information in a training example is sensitive. This problem is particularly pertinent in NLP, as NLP training data are often mixed with sparse domain-dependent private information, and not all tokens need to be protected. For example, for the sentence \emph{``My SSN is \textcolor{amber}{\hlapricot{123-45-6789}}''}, only the last few tokens of the actual SSN need to be protected. 

In fact, the definition of DP does \emph{not} prevent us \emph{at all} from protecting only the sensitive part of data. 
Specifically, DP ensures that the output of a data analysis algorithm stays roughly the same for neighboring datasets, 
while providing the flexibility to adjust the definition of neighboring relation to specific application contexts.
\citet{shi2021selective} recently proposed an instantiation of DP, called Selective-DP (SDP), which defines neighboring datasets to differ only in the sensitive part of a training example and as a result, SDP \emph{selectively} hides the difference in the sensitive part only. 
SDP is particularly suitable for NLP and many other unstructured, high-dimensional data, wherein sensitive information only accounts for a small part. 
But their privacy mechanism to achieve SDP suffers from three  problems: 1) it requires substantial knowledge 
about the model to separate the private and public variables,
and it is unclear how their algorithm tailored to recurrent neural networks could be extended to modern Transformer-based NLP models;
2) it has only been evaluated with explicit private entities but not with contextual sensitive information; 3) it doesn't provide protection for undetected sensitive tokens; These constraints limit the applicability of SDP in real-world scenarios.   

Large language models (LLMs)  \cite{vaswani2017attention}  
have achieved tremendous success in NLP. 
They are pretrained on a massive amount of public textual data,
 and thus excel at capturing general language structures. 
A common practice in NLP is to fine-tune these LLMs on downstream tasks. Such a fine-tuning process also works well in the private training context. 
Previously, \citet{yu2021differentially} showed that 
privately fine-tuning an additional small set of parameters on top of off-the-shelf LLMs with private data  achieves comparable performance to non-private baselines. 
Inspired by their findings, in this paper, we propose a two-phase fine-tuning privacy mechanism, \emph{Just fine-tune twice} (\textsc{Jft}),  to achieve SDP for LLMs. 
Instead of directly using off-the-shelf models to fine-tune once, we have two fine-tuning steps: 1) we first redact the in-domain  data of the downstream tasks, and  fine-tune the model with these in-domain redacted data  (\textit{redacted}-fine-tune), and 2) then privately fine-tune the model on the original private data (\textit{private}-fine-tune). 
This additional \textit{redacted}-fine-tune step allows the model to directly learn information from the in-domain data 
and thus leads to a better model initialization for the second \textit{private}-fine-tune step. Moreover, in the \textit{redacted}-fine-tune step, we show that even with limited public data (where manual screening is possible), \textsc{Jft} achieves better utility than fine-tune-once baselines. Additionally, we can apply lightly-noised optimizers and privacy amplification to  protect undetected sensitive tokens. 



Our contributions are as follows. First, we propose an effective and generalizable privacy mechanism 
to achieve SDP for large language models for various NLP tasks. 
Second, we design secret detectors of different privacy levels (explicit and contextual sensitive data) and study their implications on the models. 
Third, our method can utilize even a small amount of public data to achieve better utility, and mitigate the missed sensitive token problem with lightly-noised optimizer and privacy amplification. 
Finally,  we show that, opposite to the common belief that privacy is at odds with utility, private learning doesn't have to conflict with the utility because private information in the data could be irrelevant to the learning task.

\vspace{-0.5em}
\section{Preliminary}


A differentially private algorithm hides the difference between two neighboring datasets.

\begin{definition}[Differential Privacy]
Given a domain $\mathcal{D}$, any two neighboring datasets $D,
D'\subseteq \mathcal{D}$ 
a randomized algorithm $\mathcal{M}: \mathcal{D} \rightarrow \mathcal{R}$ is $(\epsilon_w, \delta_w)$-differential private if for all neighboring datasets $D$ and $D'$ and all $T \subseteq \mathcal{R}$, 
\begin{align*}
    \Pr[\mathcal{M(D)}\in T] \leq e^{\epsilon_w}\Pr[\mathcal{M(D')}\in T] + \delta_w.
\end{align*}
\end{definition}

The neighboring relation captures what is protected. Traditional DP literature has considered neighboring datasets as those differing in one training example; thus, the corresponding DP protects each training example as a whole. We denote by $\epsilon_w$ and $\delta_w$ the privacy parameters achieved under this traditional neighboring relation definition. Given the sparsity of sensitive information in language data, this instantiation of neighboring relations is apparently over-pessimistic.
\citet{shi2021selective} proposed Selective-DP (SDP), which instantiates the neighboring datasets to be those that differ in the sensitive attributes of a training sample; as a result, SDP \emph{selectively} hides the difference in the sensitive part only. In the context of NLP, a training example could be a sentence or a paragraph depending on the task and the attributes are individual tokens. 

In this paper, we will focus on designing learning algorithms to achieve SDP. Formally, SDP relies on a policy function $F$ that specifies the sensitive information in a training example to be protected in an \emph{application-dependent} fashion.

\begin{definition}[Policy Function]
A policy function $F:\tau\rightarrow \{0, 1\}^{|r|}$ decides which attributes of an example $r\in \tau$ are public ($F(r)_i=1$) or private ($F(r)_i=0$). $|r|$ is the number of attributes in $r$.
\label{def:policy function}
\end{definition}



Detecting private information manually in a large corpus based on the policy function is often costly. In that case, one may resort to building automatic \emph{secret detectors} 
to identify the sensitive attributes.  A simple example of a secret detector is a regular expression to capture phone numbers.  
However, secret detectors could miss some private attributes and produce false negatives, which intuitively would weaken the privacy guarantees. Existing work~\cite{doudalis2017one, shi2021selective,zhao2022provably} that selectively protects data either assumes a perfect detector or uses an overly conservative detector with a low false negative but at the cost of a high false positive. In this paper, we provide alternative ways to address this issue with a better privacy-utility tradeoff (Section~\ref{sec:method}).

With $F$, SDP defines $F$-Neighbors. 
\begin{definition} ($F$-Neighbors). Consider a policy function $F$ and two datasets $D$ and $D'$. $D'$ is a $F$-neighbor of $D$ (denoted by $D'\in N_F(D)$) if and only if $\exists r\in D$ s.t., 
$F(r)$ has at least one private attribute, $\exists r'\in D'$ s.t., $F(r)$ and $F(r')$ differ by at least one private attribute, and $D'=D\backslash \{r\} \cup \{r'\}$. 
\label{def:neighboring datasets}
\end{definition}
\vspace{-1.5em}

\begin{figure*}
    \centering
    \includegraphics[scale=0.45]{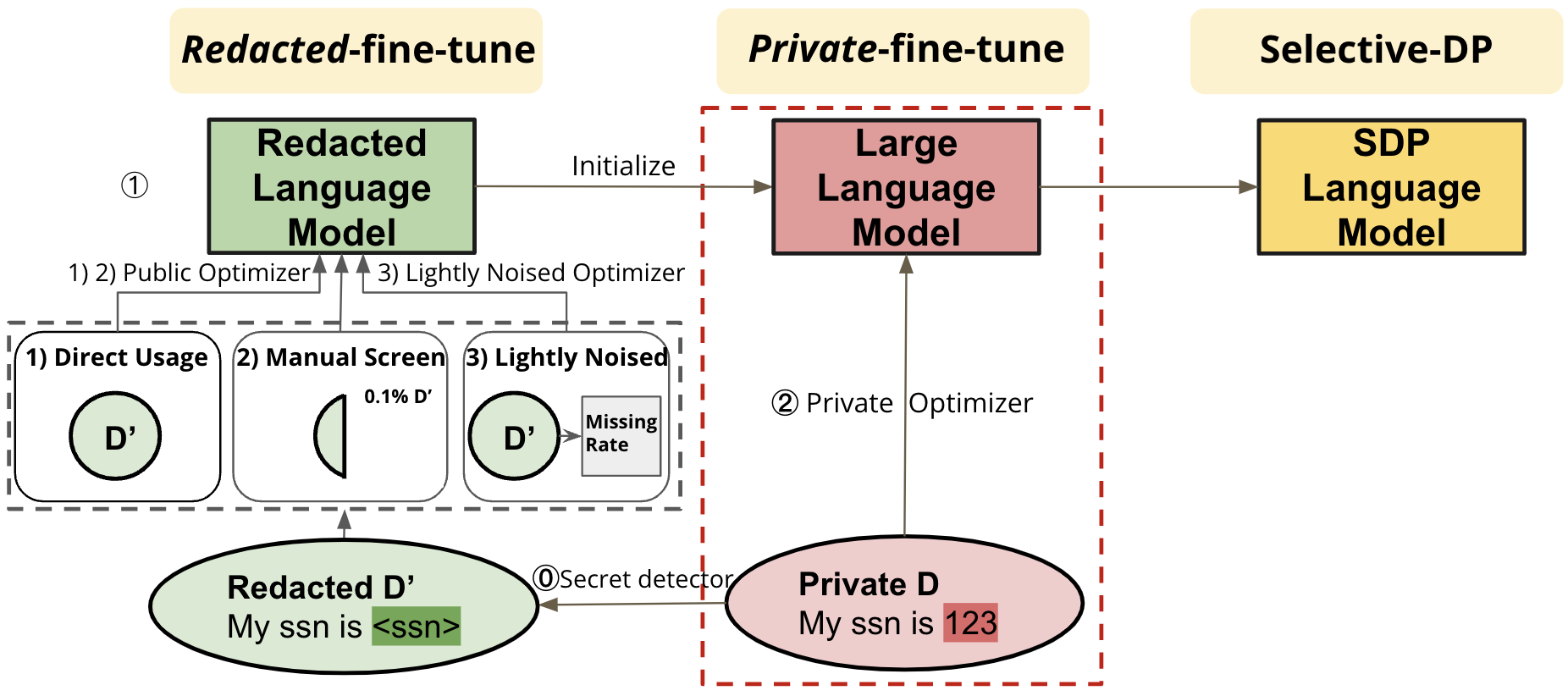}
    \caption{The two-phase \textsc{Jft} mechanism. As pre-processing, 
    we  apply the secret detector to redact  the private data $D$ and obtain the redacted data $D'$.  Next, depending on the detector's performance,  we use different ways to fine-tune the language model on the redacted $D'$  and obtain a redacted model.  Then we fine-tune the model again on the private data $D$ with a private optimizer (e.g., DPSGD) to achieve an SDP-protected model. }
    \label{fig:model}
\end{figure*}

Given the definition, the dataset with \emph{``My ID is \textcolor{amber}{\hlapricot{123}}''} and the dataset with \emph{``My ID is \textcolor{teal}{\hllightcyan{456}}''} are $F$-Neighbors (because except for the actual ID number, the other tokens are the same), while the datasets with \emph{``Hi there''} and the dataset with \emph{``Hello there''} are not $F$-neighbors because the only token they differ in is not sensitive.
An SDP algorithm guarantees that $F$-neighbors cannot be distinguished by attackers if they observed the output.
\begin{definition}
(Selective Differential Privacy). Under a policy function $F$,  
a randomized algorithm $\mathcal{M}: \mathcal{D} \rightarrow \mathcal{R}$ satisfies $(F, \epsilon_s, \delta_s)$-selective differential privacy if for $\forall D, D'\in N_F(D)$,  and  $\forall T \subseteq \mathcal{R}$,  $\Pr[\mathcal{M}(D)\in T] \leq e^{\epsilon_s}\Pr[\mathcal{M}(D')\in T] + \delta_s.$
\end{definition}

\wyshi{
In this paper, we differentiate between SDP and DP for two reasons. Firstly, we want to highlight that the privacy parameters  associated with SDP ($\epsilon_s$ and $\delta_s$) and DP ($\epsilon_w$ and $\delta_w$) are incomparable. For instance, one cannot claim which of (1, 0.001)-SDP and (2, 0.001)-DP provides stronger privacy guarantees because they are under different privacy notions. To meaningfully present the value of these privacy parameters, we need to specify under which definition these parameters are calculated; to meaningfully compare the value of these parameters, we need to make sure that the parameters are calculated under the same privacy definition. Secondly, we would like to remain the same terminology as our main reference, \citet{shi2021selective}, which also uses the terms SDP and DP to refer to the privacy guarantees under the two different neighboring relations. In the rest of the paper, we use different notations to distinguish the privacy parameters associated with SDP ($\epsilon_s$ and $\delta_s$) and DP ($\epsilon_w$ and $\delta_w$). 
}

\vspace{-2em}
\section{\textsc{Jft}: Just Fine-tune Twice}
\label{sec:method}

Now we describe \textsc{Jft}, a two-phase privacy mechanism to achieve SDP for large language models (Figure~\ref{fig:model}). 
In the first \textbf{redacted-fine-tune} phase, we redact the private data $D$ with a secret detector to obtain the redacted version $D'$,  and learn a redacted model from $D'$ in a privacy-preserving way. 
 In the second \textbf{private-fine-tune} phase, we further fine-tune the redacted model (from phase one) on the private data $D$ with a private optimizer to achieve SDP guarantees.


\subsection{Phase 1: \textit{Redacted}-fine-tune}
\label{sec: phase 1}
\tool is built upon the observation that the public portion of the in-domain data does not require protection 
and can be utilized in various ways to help the model learn in-domain information. 
In this phase, we apply the secret detector to redact the private data $D$ and obtain redacted in-domain data $D'$. Dependent on the detector performance, we propose the three following methods to use $D'$ to fine-tune off-the-shelf language models.


\noindent\textbf{Direct Usage.} If the secret detector masks all the sensitive information in $D$ (which is possible when $D$ is small enough to support thorough inspection or when a detector is very conservative and removes most of the essential information, see examples in Table~\ref{tab:policy function results}), we can use the redacted $D'$ directly to fine-tune the model with a public and unnoised optimizer like SGD. 

\noindent\textbf{Selective Manual Screening.} 
If the secret detector is imperfect, we can select an affordable subset from $D'$ and then manually sanitize all the missed secrets. 
Then we fine-tune the model with this small sanitized subset with a public optimizer.  Experiments show that even with a small amount of sanitized in-domain data, the resulting model still outperforms the traditional DP learning algorithms that pessimistically protect every single token. 

\noindent\textbf{Lightly-Noised Fine-tuning.} 
When the detector is imperfect, besides manually screening out the missed secrets, we could also employ a private optimizer to train on $D'$ that contains missed sensitive tokens. Because missed tokens only account for a small portion of $D'$, intuitively, a much smaller noise is needed to ensure the privacy of the missed tokens than the noise magnitude required to ensure the privacy of the entire $D'$. We propose to leverage \emph{privacy amplification by subsampling} (PAS)~\cite{balle2018privacy} to calculate the privacy parameters associated with the private optimizer. 
The intuition of PAS is that if we perform a DP mechanism on random subsamples of the data, and  one data point is not included in the subsamples, nothing about it could be leaked. In this way, we could amplify the privacy guarantee. In our scenario, we need to protect the missed sensitive tokens. If we know the secret detector's missing rate $m$ (i.e., $m$=number of missed sensitive tokens/total tokens, the probability of sampling a missed secret), we can calculate the privacy budget $\epsilon_s$ by privacy amplification using the subsampling ratio $m$. 

Note that the application of PAS requires the number of missed tokens that appear in any batch to be the same, which does not necessarily hold. Hence, the privacy parameters calculated from privacy amplification are an empirical estimate of the actual privacy loss. 
In practice, the secret detector's missing rate is unknown and we need to estimate it (denoted as $\widetilde{m}$). Then we change the original sampling rate $p_0$ in moment accounting-based privacy parameter calculation~\cite{abadi2016deep} to $p=p_0*\widetilde{m}$ and calculate the noise injected into each private optimizer iteration according to a predefined privacy budget $\epsilon$ under $p$.  


In our experiments, we sample 0.01\% training data for 10 times, and estimate the 95\% confidence interval of the missing rate, [$\widetilde{m}_{\text{low}}$, $\widetilde{m}_{\text{high}}$]. For both $\widetilde{m}_{\text{low}}$ and $\widetilde{m}_{\text{high}}$, we can calculate an associated $\epsilon_{\text{low}}$ and $\epsilon_{\text{high}}$ according to Theorem 9 in \citet{balle2018privacy}, and report both $\epsilon$.


\subsection{Phase 2: \textit{Private}-fine-tune.}  

In the second phase, we initialize the model with the  redacted model from phase one, and fine-tune it with the original private $D$ and a private optimizer (e.g., DPSGD \cite{abadi2016deep} or any other more advanced private optimizer that achieves DP).   

Unlike the privacy mechanism in \citet{shi2021selective}, our algorithm does not require  knowledge about the models or the tasks, and therefore can be easily applied to different models such as GPT2 \cite{radford2019language} and Roberta \cite{liu2019roberta}, and different tasks such as language generation and natural language understanding. See Section~\ref{sec:implementation detail} for more implementation details. 

\noindent\textbf{One-phase vs two-phase.} Compared to conventional differentially private training, our algorithm introduces an additional stage that involves redaction and regular unnoised training on redacted data. In fact, the computational cost of the additional stage is much lower than the cost originally incurred by DP learning, because the additional first phase does not need costly per-sample gradient clipping and noising operations. And redaction is a  common first-step people are already doing and familiar with. Also, there exist abundant off-the-shelf tools that allow redaction at scale.

\subsection{Privacy Analysis}
\wyshi{We provide Theorem~\ref{thm:sdp} for privacy analysis. It ensures that, if the user has a secret detector with 100\% recall, JFT-trained models achieve (0,0)-SDP after phase one and ($\epsilon$, $\delta$)-SDP after phase two. A secret detector with 100\% recall is possible if the user can afford manual inspection or have enough domain knowledge. When a detector with 100\% recall is not  possible, we use \emph{lightly-noised fine-tuning} to empirically protect the missed secrets as mentioned in Section~\ref{sec: phase 1}. }




\begin{theorem}
\label{thm:sdp}
Given that 1) in the first phase, the data used for fine-tuning do not contain sensitive tokens and a public optimizer is used, and 2) in the second phase, the private optimizer achieves $(\epsilon,\delta)$-DP, \tool achieves $(\epsilon, \delta)$-SDP.
\end{theorem}

The proofs are deferred to Section~\ref{sec:proof}. The theorem shows that under direct usage or selective screening of $D'$, \tool achieves SDP with the same privacy parameter values as the ones pertaining to the private optimizer used in the second phase. 

\section{Secret Detectors of Different Levels}
\label{sec:secret detector}

Typical private information includes personal-identifiable information (PII) 
such as name and birthday. 
But as pointed out in \citet{brown2022does}, 
one key challenge in NLP is that private information is often 
contextual. 
For example, they presented a dialogue  between Alice and Bob about Alice's divorce (Table~\ref{tab:policy function results}): none of the tokens in \emph{``What are you going to do about the custody of the kids?''},  are PII by themselves, 
but combined together, the semantics reveals private information.

To build generalizable secret detectors, we utilize off-the-shelf NER, dependency parser, and POS tagger in spaCy \cite{spacy2} to label each token, and redact different sets of tokens to achieve the different privacy levels below (entity level and contextual level).  
To qualitatively show their protection levels, we apply them to redact two sentences from the divorce dialogue in \citet{brown2022does}. The results are in Table~\ref{tab:policy function results}.  




\begin{table}[htb!]
\small
\centering
\begin{adjustbox}{width=.98\columnwidth}
\begin{tabular}{p{0.18\linewidth} | p{0.5\linewidth}|p{0.4\linewidth}}
\toprule
\small
\textbf{Secret Detector} & \textbf{What are you going to do about the custody of the kids?}                                                                                                                             & \textbf{Did you hear Alice is getting divorced?}                                                                                    \\
\midrule
Low entity            & What are you going to do about the custody of the kids?                                                                                                                                      & Did you hear $<$PERSON$>$ is getting divorced??                                                                    \\
\midrule
High entity              & What are you going to do about the custody of the kids?                                                                                                                                      & Did you hear $<$PERSON$>$ is getting divorced??                                                                    \\
\midrule
Low contextual        & $<$PRON$>$ are $<$PRON$>$ going to do about $<$OBJ$>$ of the $<$OBJ$>${}?                                                & Did $<$PRON$>$ hear $<$PROPN$>$ is getting divorced??                                             \\
\midrule
High contextual          & $<$PRON$>$ are $<$PRON$>$ $<$VERB$>$ to $<$VERB$>$ about $<$OBJ$>$ of the $<$OBJ$>${}? & Did $<$PRON$>$ $<$VERB$>$ $<$PROPN$>$ is getting $<$VERB$>${}??\\
\bottomrule
\end{tabular}
\end{adjustbox}
\caption{Results of different secret detectors on the two example sentences from \citet{brown2022does}.
}
\label{tab:policy function results}
\end{table}

\noindent\textbf{Low entity} redacts four types of named entities (person, organization, date, and location), which are considered PII defined by the US Department of Labor\footnote{\url{https://www.dol.gov/general/ppii}}. We use NER in spaCy to detect them. If we apply this detector, \textit{``Did you hear Alice is getting divorced?''} becomes \textit{``Did you hear $<$PERSON$>$ is getting divorced?''} An attacker who attacks a model trained on the latter sentence can at best learn about the divorce but cannot know who.

\noindent\textbf{High entity} redacts all the 18 entities in spaCy including the four above and more, like time entity\footnote{For the full list, see \url{https://spacy.io/usage/linguistic-features\#named-entities}.}.

The two secret detectors above rely on named entities, so they are more explicit than the two detectors below, which consider the overall sentence structure and thus are more contextual.  

\noindent\textbf{Low contextual} protects all the 18 entities plus proper nouns, pronouns, and sentence subjects and objects. 
This detector drastically increases the privacy level:  
we cannot get any useful information from the left example in Table~\ref{tab:policy function results}. 

\noindent\textbf{High contextual} 
further redacts all the verbs, in addition to the tokens redacted by the low contextual detector. It increases the privacy level even further and we cannot learn anything from both examples. This is to \emph{stress-test} \tool and see the model utility when the majority of the tokens are redacted 
\footnote{Note that ``divorced'' is actually an adjective in the example. As mentioned earlier, we can address the detector's mistakes with lightly-noised fine-tuning. 
}.

Human language is diverse, and private information can take various forms. 
So instead of designing sophisticated algorithms, we intentionally rely on common NLP tools to build easy-to-use domain-agnostic secret detectors with  \emph{high recalls} to protect privacy as much as possible. 
As shown in Table~\ref{tab:policy function results}, these detectors tend to over-sanitize the sentences.   But we will show later, even with over-redaction, \tool still achieve good performance. Private textual information can be treated more sophisticatedly, but how to better detect private information is not the focus of this paper. Our goal is to show that simply redacting tokens achieves promising performance and \tool is compatible with better private information detection algorithms to further improve the results. 




\section{Experiments}
We conduct experiments on two NLP tasks: \emph{1)} natural language understanding (NLU, on GLUE) and \emph{2)} language generation (on Wikitext-2 and ABCD). 

\noindent\textbf{Datasets.} \emph{1)} \textbf{GLUE} \cite{wang2018glue} is a widely-used multi-task benchmark dataset for NLU. It contains sensitive information such as name and date. \emph{2)} \textbf{Wikitext-2} \cite{merity2016pointer} contains Wikipedia articles  with private information such as name and date. 
\emph{3)} \textbf{ABCD} \cite{chen2021action} is a human-human customer service dialogue dataset under real-world scenarios with user private information such as name and order IDs.

\noindent\textbf{Models.} We use Roberta \cite{liu2019roberta} for the NLU classification task and GPT2 \cite{radford2019language} for the language generation task.  
Due to computational constraints, we use Roberta-base and GPT2-small for the experiments. We use an efficient implementation of DPSGD in \citet{li2021large}.  
Based on previous studies \cite{li2021large, yu2021differentially}, larger DP models usually achieve better results and thus we expect that larger SDP models will achieve even better performances.

\noindent\textbf{Baselines.} \emph{1)} \textbf{No-DP}: the model is fine-tuned using regular Adam optimizer \cite{kingma2014adam} without extra noise and hence it does not have any privacy guarantees (i.e., $\epsilon_w=\epsilon_s = \infty$). \emph{2)} \textbf{DPSGD}\footnote{\wyshi{The public baseline from Roberta does not use infilling, for a fair comparison, we chose the DPSGD baseline without infilling from \cite{li2021large}. But 
compared to DPSGD baselines with infilling from \cite{li2021large}, JFT without infilling with a conservative secret detector is still better or comparable. 
Previous studies found that infilling improves the results, so it is possible that with infilling, \tool can achieve even better results. }
}: the model is fine-tuned with traditional DPSGD~\citet{abadi2016deep} where the gradient is clipped and noised in every gradient descent iteration (we employ the DP-Adam variant where the optimizer is Adam but its gradient privatization is the same as DPSGD, and we keep the term DPSGD as it is more accessible to the community). While DPSGD was originally proposed to achieve the DP guarantees that protect a training example as a whole, it can also achieve SDP guarantees with the same privacy parameters (i.e., $\epsilon_s=\epsilon_w$ and $\delta_s=\delta_w$). \emph{3)} \textbf{CRT}: the model is trained with 
the recently proposed Confidentially Redacted Training \cite{zhao2022provably} that achieves $(\epsilon_c, \delta_c)$-Confidentiality. Confidentiality is a new definition related to SDP but different from SDP, it ensures the indistinguishability between a secret and a <MASK> token, so its privacy parameters are not directly comparable to SDP. 
Thus, we add the same amount of noise to CRT and SDP, empirically compare SDP and CRT with the canary insertion attack in Figure~\ref{fig:canary} and~\ref{fig:canary_1}, and report the utility in Table~\ref{tab:CRT vs SDP} in the Appendix. 
\emph{4)} \textbf{Redacted}: We also present the utility of the redacted models since they are also privacy-preserving. Note when the secret detector is perfect, the redacted models have a perfect SDP privacy guarantee (i.e., $\epsilon_s=0$). However, it does not allow the model to learn from sensitive tokens at all. \textsc{Jft}, by contrast, empowers the model to learn from sensitive data with a flexible, tunable tradeoff between privacy and utility. Moreover, it provides ways to offer quantifiable privacy in the presence of imperfect secret detectors. 

\noindent\textbf{Our models.} \emph{1)} \textbf{\tool}: this is our \tool model directly using the redacted data in phase one. \emph{2)} \textbf{\tool+manual screening}: this is \tool using a subset of the redacted data where missed secrets are manually filtered out in phase one. \emph{3)} \textbf{\tool+light noise}: this is \tool  where we add light noises according to the estimated missing rate in phase one. 

\begin{table*}[!htbp]
    \centering
    \resizebox{\textwidth}{!}{
    \begin{tabular}{l|l|l|cc|l|cc|l|cc|l|cc||l|cc|l|cc}
    \toprule
   \multicolumn{2}{l|}{\textbf{Direct Usage}}&\multicolumn{12}{c||}{\textbf{NLU on GLUE, $\delta_s$=1/2$|D_{\text{train}}|$}} & \multicolumn{6}{c}{\textbf{Language Generation, $\delta_s$=1e-6}}\\
    \midrule
   \multicolumn{2}{l|}{\textbf{}}&\multicolumn{3}{c|}{\textbf{\textsc{MNLI}}} & \multicolumn{3}{c|}{\textbf{\textsc{QQP}}}&
   \multicolumn{3}{c|}{\textbf{\textsc{QNLI}}}&
   \multicolumn{3}{c||}{\textbf{\textsc{SST-2}}}&
   \multicolumn{3}{c|}{\textbf{\textsc{WikiText-2}}}&
   \multicolumn{3}{c}{\textbf{\textsc{ABCD}}}
   \\
\midrule
 \textbf{Model}
&\textbf{Detector}& \textbf{Pct} & \textbf{Acc$\uparrow$  } 
&  \textbf{$\epsilon_s$} &   
\textbf{{Pct}} & \textbf{Acc$\uparrow$} 
&  \textbf{$\epsilon_s$} &
\textbf{Pct} & \textbf{Acc$\uparrow$} 
&  \textbf{$\epsilon_s$} &
\textbf{Pct} & \textbf{Acc$\uparrow$} 
&  \textbf{$\epsilon_s$} &
\textbf{Pct} & \textbf{PPL$\downarrow$  } 
&  \textbf{$\epsilon_s$} &
\textbf{Pct} & \textbf{PPL$\downarrow$  } 
&  \textbf{$\epsilon_s$} 
 \\ 
 \midrule
No-fine-tune&-&-&31.82 & -  
&-&36.82 &- &-&50.54 &-&-& 50.92&- &-&30.08 & -  
&-  &13.60 &-\\
\midrule
No-DP 
&-&-&87.60 & -  
&-&91.90 &- &-&92.80 &-&-& 94.80&- &-&20.48 & -  
&-  &4.96 &- 

\\

DPSGD 
&-&-& 82.10 & 2.75
& -   &85.41&2.75 
& - & 84.62&2.57
& - &86.12&2.41

&-& 27.05 & 2.58
& -& 8.31 & 2.65 

  \\
DPSGD (+spe) 
&-&-& - & -
& -   &-&- 
& - & -&-
& - &-&-

&-& 30.32 & 2.58
& -& 17.75 & 2.71 

  \\
\midrule


Redacted & low ent & 6.09\% & 86.67 & - & 6.05\% & 88.74 & - & 12.19\% & 89.64 & - & 1.79\% & 93.58 & - & 11.3\% & 22.50 & - & 2.7\% & 6.98 & -\\
\textsc{Jft} & low ent & 6.09\% & \textbf{85.74} & 0.92 & 6.05\% & \textbf{88.19} & 2.58 & 12.19\% & \textbf{89.57} & 2.37 & 1.79\% & \textbf{92.09} & 2.06
& 11.3\% & \textbf{21.86} & 2.58 & 2.7\% & \textbf{6.09} & 2.71
\\

\midrule


Redacted & high ent & 8.63\% & 86.50 & - & 8.30\% & 88.36 & - & 17.18\% & 88.96 & - & 3.01\% & 93.58 & -& 16.4\% & 24.32 & - & 3.1\% & 7.32 & -\\
\textsc{Jft} & high ent & 8.63\% & \textbf{85.61} & 0.99 & 8.30\% & \textbf{88.05} & 2.58 & 17.18\% & \textbf{89.35} & 2.37 & 3.01\% & \textbf{92.20} & 2.12
& 16.4\% & \textbf{22.55} & 2.58 & 3.1\% & \textbf{6.25} & 2.71
\\
\midrule

Redacted & low ctx & 31.19\% & 85.14 & - & 32.61\% & 85.59 & - & 35.68\% & 85.30 & - & 22.19\% & 92.55 & -& 34.8\% & 37.90 & - & 22.3\% & 28.28 & -
\\
\textsc{Jft} & low ctx & 31.19\% & \textbf{85.02} & 1.23 & 32.61\% & \textbf{87.00} & 2.41 & 35.68\% & \textbf{87.99} & 2.52 & 22.19\% & \textbf{92.43} & 2.17& 34.8\% & \textbf{25.62} & 2.58 & 22.3\% & 8.80 & 2.71
\\
\midrule
\multicolumn{14}{l}{\textbf{Stress-test}}        \\        
\midrule

Redacted & high ctx & 44.27\% & 83.23 & - & 45.93\% & 83.48 & - & 45.59\% & 82.81 & - & 38.13\% & 91.86 & -& 45.0\% & 54.29 & - & 28.6\% & 65.45 & -
\\
\textsc{Jft} & high ctx & 44.27\% & \textbf{84.11} & 1.18 & 45.93\% & \textbf{86.42} & 2.67 & 45.59\% & \textbf{87.06} & 2.41 & 38.13\% & \textbf{91.17} & 2.17& 45.0\% & 27.19 & 1.96 & 28.6\% & 12.93 & 2.71\\

\bottomrule

\end{tabular}
}
    \caption{Model utility and privacy guarantee on GLUE dev sets for NLU (left) and WikiText-2 and ABCD for generation (right). \textbf{Detector}: the secret detector to realize the policy function. \textbf{low ent}: low entity detector. \textbf{low ctx}: low contextual detector.  \textbf{Pct}: the percentage of sensitive tokens.   \textbf{Acc}: accuracy. \textbf{PPL}: perplexity. \textbf{$\epsilon_s$}: SDP privacy guarantee. \textbf{DPSGD(+spe)}: DPSGD with added special tokens.  For QNLI and SST-2, $\delta\approx$1e-5; for MNLI and QQP,  $\delta\approx$1e-6 due to the large data size; No-DP and DPSGD  results are from \citet{liu2019roberta} and \citet{li2021large}. For generation tasks,  $\delta$=1e-6, we train the baselines and report the results.  
    The models in bold are better than DPSGD. 
    }
    \label{tab:glue utility}
\end{table*}

\section{Results}

We show three major findings: \emph{1)}  the impacts of secret detectors are task-dependent on the resulting \textsc{Jft} models, but even for conservative contextual detectors (30\%+ tokens are redacted), \textsc{Jft} still achieves better results than naive DPSGD (Section~\ref{sec:result,different secret detectors}); 
\emph{2)} despite the small scale, using the manual screened in-domain data still improves the \textsc{Jft} model utility (Section~\ref{sec:result,manual screen}); 
\emph{3)} lightly noised optimizer with privacy amplification protects missed sensitive tokens from attacks (Section~\ref{sec:result,privacy amplification}). 

\wyshi{There is always a privacy-utility trade-off, so larger epsilons lead to better utilities but worse privacy. And when comparing models, we need to look at the model utility under a similar privacy budget. 
An epsilon of 1 to 3 is commonly used in various privacy literature \cite{yu2021differentially, li2021large, zhao2022provably}. In our experiments, we  pre-calculated the privacy parameter so that an $\epsilon$ of around 3 is spent when training ends. }


\subsection{Secret Detectors of Different Levels}
\label{sec:result,different secret detectors}

Table~\ref{tab:glue utility} 
show the results on GLUE (left) and generation (right). 
$Pct$ is the percentage  of sensitive tokens redacted by the detector. 
$\epsilon_s$ is the SDP privacy budget, the lower the better. 
We compare model utility under a similar privacy budget $\epsilon_s$.


\noindent\textbf{Natural Language Understanding.} Table~\ref{tab:glue utility} (left) shows that under a similar $\epsilon_s$, 
all the \textsc{Jft} models achieve better performance than the DPSGD baseline, 
even  when over 40\% of tokens 
are redacted. 

Besides, for all the tasks, 
all the redacted models achieve reasonable utility, 
even when a large portion of the tokens are redacted. For example,   
the redacted model (high contextual) is better than DPSGD on MNLI ($83.23$ vs $82.10$, $44.27$\% redacted) and  SST-2 ($91.17$ vs $86.12$, $38.13$\% redacted). 
This confirms the motivation of SDP that when building private NLP models, we should not naively protect all the tokens regardless of their properties. Instead,  we should consider if the sensitive tokens will impact the task. If not, we can simply redact them to build private models.

 
Also, if \textsc{Jft} can improve the redacted model depends on the task. For SST-2 on sentiment analysis, 
the \textit{private}-fine-tune step does not improve the redacted model. 
This is because, the redacted models achieve a high accuracy (even the worst accuracy is $91.86$, only a $2.94$ drop from the SOTA public model with an accuracy of $94.8$), 
and fine-tuning them on the private data with noisy gradients is not enough to close the small gap.  
But for tasks with a bigger gap between the redacted and No-DP models (e.g., MNLI, QQP, and QNLI), \textsc{Jft} can further improve the redacted model. 
Besides, the gap between the  redacted model and the corresponding \textsc{Jft} model becomes bigger as the privacy level increases: for QNLI (low contextual), the gap is $87.99$-$85.30$=$2.69$, while for QNLI (high contextual), 
the gap is $87.06$-$82.81$=$4.25$. This shows the model does learn useful information from the sensitive tokens during the  \textit{private}-fine-tune step. 



\noindent\textbf{Language Generation.} Table~\ref{tab:glue utility} (right) shows the language generation results. 
We  note that language generation is different from NLU tasks, because for NLU, the models used for initialization without any fine-tuning (``No-fine-tune'' in Table~\ref{tab:glue utility}) start with a bad accuracy ($\leq$ $50$\%, just random guess), and adding special tokens to it would still start with a random guess, so additional special tokens  will not impact the final results greatly. But for the generation task, the ``No-fine-tune'' GPT2 is already a strong  model for initialization with ppl=$30.08$ on Wikitext-2 and $13.60$ on ABCD, and we found that adding special tokens would disturb this initialization and greatly impact the final result. Because all the \textsc{Jft} models have added special  tokens like ``<MASK>'' and ``SYS:'',  for a fair comparison, we report two DPSGD baselines, one without special tokens (``DPSGD'') and one with special tokens (``DPSGD (+spe)'').  See Section~\ref{sec:implementation detail} for  more discussions on the impact of special tokens.  



Compared to ``DPSGD (+spe)'', all \textsc{Jft} models achieve better model utility on both datasets. For ``DPSGD'' without special tokens, fine-tuning on the downstream tasks  improves the model from $30.08$ to $27.05$ (the improvement $\Delta$=$3.03$); for \textsc{Jft} (low contextual), it is initialized with the redacted model with   ppl=$37.90$, and privately fine-tuning it improves the perplexity  to $25.62$ ($\Delta$=$12.28$). This shows that although the initialization seems worse ($30.08$ vs $37.90$), since the redacted model is fine-tuned directly on  in-domain redacted data, it does learn useful information from the first \textit{redacted}-fine-tune step, 
and the second \textit{private}-fine-tune step can further improve upon the redacted model. For \textsc{Jft} (high contextual) for the stress test, although $45$\% tokens  are masked and the language structure is largely impacted, \textsc{Jft} still improves the redacted model utility from $54.29$ to $27.19$ ($\Delta$=$27.10$) and performs on par with  DPSGD ($27.05$ vs $27.19$). 




\subsection{Selective Manual Screening}\label{sec:result,manual screen}

\begin{table*}[!htbp]
    \centering
    \resizebox{\textwidth}{!}{
\begin{tabular}{l|cc|cc|cc|cc||cc|cc}
\toprule
                    & \multicolumn{2}{c|}{\textbf{MNLI}}              & \multicolumn{2}{c|}{\textbf{QQP}}               & \multicolumn{2}{c|}{\textbf{QNLI}}              & \multicolumn{2}{c||}{\textbf{SST-2}}       & \multicolumn{2}{c|}{\textbf{WikiText-2}}        & \multicolumn{2}{c}{\textbf{ABCD}}             \\
\midrule
\textbf{Model}               & \textbf{Acc$\uparrow$}   &   95\%-$\epsilon_s$            & \textbf{Acc$\uparrow$}   &    95\%-$\epsilon_s$                      & \textbf{Acc$\uparrow$}   &     95\%-$\epsilon_s$                    & \textbf{Acc$\uparrow$}   &   95\%-$\epsilon_s$                 & \textbf{PPL$\downarrow$}   &     95\%-$\epsilon_s$              & \textbf{PPL$\downarrow$}  &     95\%-$\epsilon_s$    \\
\midrule

DPSGD               &  82.10 & 2.75                   & 85.41 & 2.75                      & 84.62 & 2.57                    & 86.12 & 2.41                & 27.05 & 2.58                       & 8.31 & 2.65           \\
\midrule
 Missing rate $m$ (95\% CI) & \multicolumn{2}{c|}{(0.3\%, 1.2\%)} & \multicolumn{2}{c|}{(0.3\%, 1.2\%)} & \multicolumn{2}{c|}{(0.1\%, 0.6\%)} & \multicolumn{2}{c||}{(0\%, 1.8\%)} & \multicolumn{2}{c|}{(0.4\%, 0.7\%)} & \multicolumn{2}{c}{(0.1\%, 1.2\%)} \\
Recall (95\% CI) & \multicolumn{2}{c|}{(87.5, 96.7)} & \multicolumn{2}{c|}{(85.9, 96.1)} & \multicolumn{2}{c|}{(96.4, 99.3)} & \multicolumn{2}{c||}{(40.2, 100)} & \multicolumn{2}{c|}{(95.6, 97.8)} & \multicolumn{2}{c}{(62.7, 97.4)} \\
\textsc{Jft}+light noise       & \textbf{82.76} & (0.08, 0.43)  & 85.28 & (1.40, 1.71)    & \textbf{84.88} & (2.29, 2.68)   & \textbf{89.33} & (0, 0.43) &  \textbf{25.21} & (2.73, 2.92)  & \textbf{5.78} & (1.08, 1.60) \\
\midrule
\multicolumn{13}{l}{\textbf{Conservative Estimation}}\\
\midrule
Missing rate $m$  & \multicolumn{2}{c|}{8.6\%} & \multicolumn{2}{c|}{8.3\%} & \multicolumn{2}{c|}{17.2\%} & \multicolumn{2}{c||}{3.0\%} & \multicolumn{2}{c|}{16.4\%} & \multicolumn{2}{c}{3.0\%} \\
Recall & \multicolumn{2}{c|}{0} & \multicolumn{2}{c|}{0} & \multicolumn{2}{c|}{0} & \multicolumn{2}{c||}{0} & \multicolumn{2}{c|}{0} & \multicolumn{2}{c}{0} \\
\textsc{Jft}+light  conservative noise &   82.00 & 0.45          &   84.77 & 2.91          &   84.02 & 2.95                   & \textbf{89.22} & 0.43     &   \textbf{26.59} & 3.03          &   \textbf{6.64} & 1.67    \\
\bottomrule
\end{tabular}
}
\caption{Privacy-amplified \textsc{Jft} performance on all the tasks, with the high entity detector. We report the estimated 95\% confidence interval (CI) of the missing rate $m$, recall and the corresponding 95\% CI of the $\epsilon_s$.  
    }
    \label{tab:privacy amplification}
\end{table*}

\begin{table}[htbp!]
\centering
\begin{adjustbox}{width=\columnwidth}
\begin{tabular}{l|cccc||cc}
\toprule

\textbf{Manual Screening}& \multicolumn{6}{c}{\textbf{$D'$ (redacted)=0.1\%$D_0$, $D$ (private)=100\%$D_0$}}                                           \\
\midrule

\multirow{2}{*}{\textbf{Task}}  
& \textbf{MNLI} & \textbf{QQP}   & \textbf{QNLI} & \textbf{SST-2} & \textbf{WikiText-2} & \textbf{ABCD} \\


& Acc $\uparrow$ & Acc $\uparrow$ & Acc $\uparrow$ & Acc $\uparrow$ & PPL $\downarrow$ & PPL $\downarrow$ \\
 \midrule
\textbf{$D'$ size} 
& 300 & 300   & 100 & 100 & 10 & 10 \\

 \midrule

\textbf{DPSGD}
& 82.10 & 85.41 & 84.62 &  86.12  & 27.05         & 8.31                                                                                                 \\
\midrule
\textbf{Redacted} 
&  52.52 & 75.25 & 66.48 & 88.88 & 28.06 & 9.36                                                                                       \\
\textbf{\textsc{Jft}+manual screening}  
&\textbf{82.45}  &\textbf{86.24} &\textbf{85.00} & \textbf{90.83} & \textbf{26.72}& \textbf{7.84}                                                                 \\
\bottomrule
\end{tabular}
\end{adjustbox}
\caption{Manual screening results on the high entity secret detector. $D_0$: original data. $D'$: the inspected redacted data. $D$: the private data. $\epsilon \approx 3$. $D'$ size is the number of records used in the \textit{redacted}-fine-tune phase. }
\label{tab:low-resource}
\end{table}

As mentioned earlier, secret detectors can miss certain secrets and we can manually filter out the missed secrets at a small scale and fine-tune with the small manually sanitized set. 
Denote the original data as $D_0$. We sample 0.1\% data from $D_0$, apply the high entity secret detectors (because it is less conservative and could miss  secrets), and manually sanitize the missed secrets to get $D'$. We use $D'=0.1\% D_0$ during \textit{redacted}-fine-tune to train the redacted model, and the entire $D_0$ as $D$ during \textit{private}-fine-tune to obtain \textsc{Jft} models. 

Table~\ref{tab:low-resource} shows the results. 
$D'$=0.1\%$D_0$ contains 100$\sim$300 examples for GLUE and 10 articles and 10 dialogues for Wikitext-2 and ABCD respectively. On all the tasks, \tool achieves better utilities than DPSGD. This shows that even fine-tuning with a small manually-screened in-domain subset can still help the model learn in-domain information, and lead to better utility. We also simulate a completely low-resource setting where we simply have limited training data (i.e., $D'$=0.1\% $D_0$, $D$=0.1\% $D_0$). See Section~\ref{sec:low-resource} for the results.


\subsection{Lightly Noised Optimizer with Privacy Amplification}
\label{sec:result,privacy amplification}
Besides manually inspecting the missed secrets, we can also use noised optimizers in the first phase to
protect the missed secrets from attacks and then adopt privacy amplification to estimate the corresponding privacy parameters. We again perform the experiments on the high entity detector and Table~\ref{tab:privacy amplification} shows the results. We convert the missing rate $m$ (\%) to the recall of the secret detector, i.e., recall=(1-$m/Pct$), where $Pct$ is the percentage of sensitive tokens among all tokens.  
``\textsc{Jft}+light noise'' shows the model performance with a noised optimizer and privacy amplification employed in the first step.  
Besides, we add more noise than needed to obtain a conservative model (``\textsc{Jft}+light  conservative noise''): for instance, MNLI data contains $8.63\%$ sensitive tokens, although the secret detector's missing rate $m$ ranges from $(0.3\%, 1.2\%)$ with $95\%$ probability, we assume $m$ to be $8.63\%$ (i.e., it miss all sensitive tokens and thus recall=$0$) to calculate and add the noises that are more than actually needed.  

The results show that  ``\textsc{Jft}+light noise''  achieves better utility than DPSGD, especially on generation tasks. The perplexity is $(25.21$ vs $27.05$, $5.78$ vs $8.31)$, and the estimated $\epsilon_s$ ranges from $(2.73,2.92)$ and $(1.08,1.60)$ with 95\% probability. Even for the conservative models, ``\textsc{Jft}+light  conservative noise'' is still better than DPSGD ($26.59$ vs $27.05$, and  $6.64$ vs $8.31$).

\subsection{Attack Results}
\begin{figure}[!htbp]
    \centering
    \includegraphics[scale=0.3]{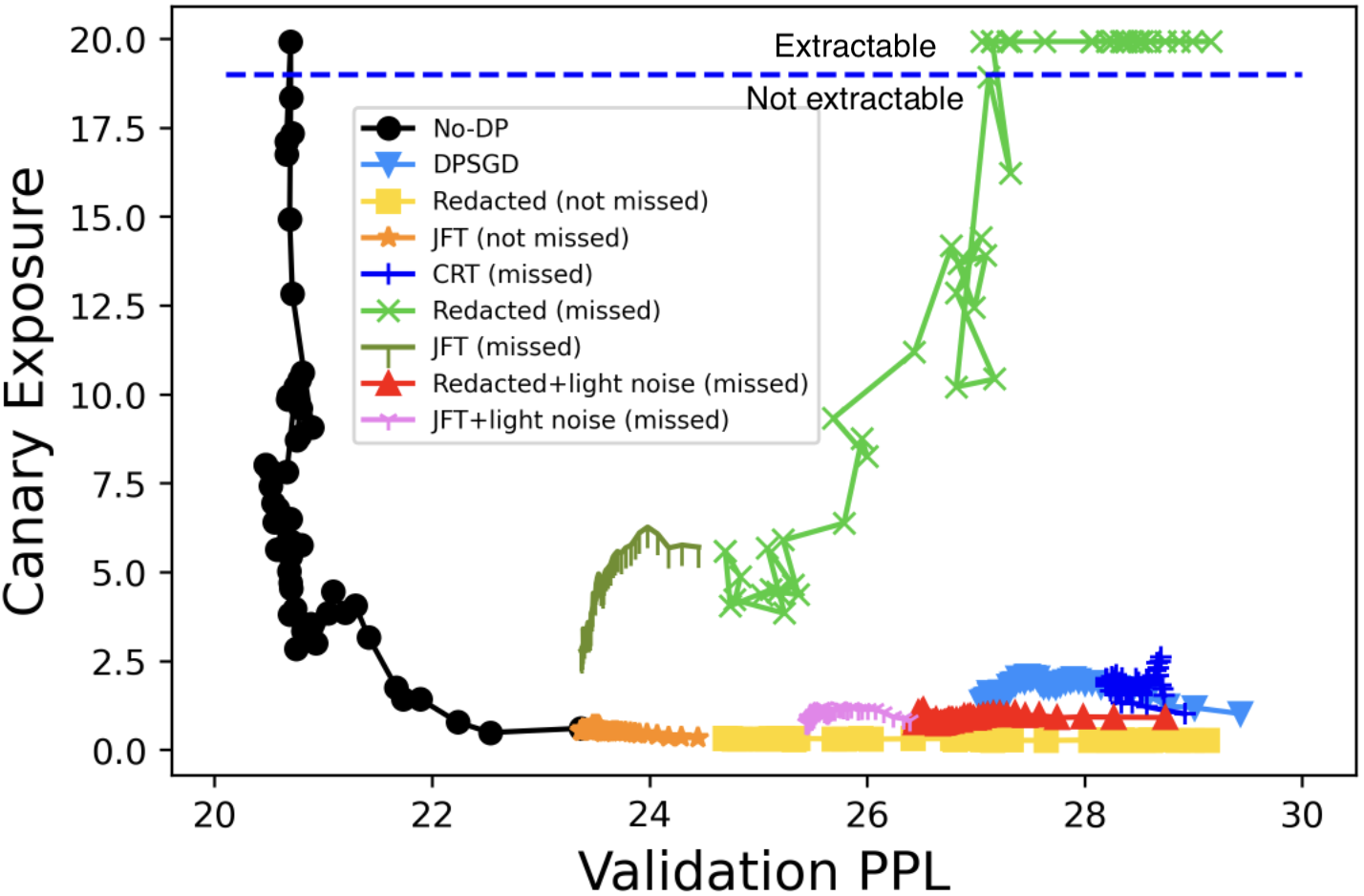}
    \caption{Canary exposure for different models. }
    \label{fig:canary}
\end{figure}

We perform the canary insertion attack \cite{carlini2019secret} to empirically show how much the models memorize the training data unintentionally. The attack is to insert a canary of a certain format into the training data, and calculate its exposure, which is the rank of the inserted canary amongst all possible values of the same format. The lower the exposure is, the safer the model is. In our experiments, we insert the canary \emph{``My ID is \textcolor{amber}{\hlapricot{341752}}''} into the training data for 10 times to make the performance difference of different models more salient. By definition, for a six-digit canary, an exposure close to $\log_2(10^{6})\approx 19.9$ means the canary can be extracted by the attackers.  The result is in Figure~\ref{fig:canary}. Each point on the figure is a model checkpoint. The X-axis shows the perplexity (utility), and the y-axis is the exposure (privacy), so Figure~\ref{fig:canary} shows different models’ privacy-utility tradeoffs.

One major reason why the model remembers a canary is that it has seen the canary many times. 
For ``No-DP'', initially, its exposure is low because it hasn't seen the canary many times. But because the model is unprotected, its exposure is unbounded and increases dramatically after it accesses the data for more epochs. This suggests that models without protection do  memorize the data unintentionally. 

For protected models (DPSGD, redacted, and \textsc{Jft}), if the canary is captured by the detector (``not missed'' in the figure), then the exposure does not increase much even if the data are accessed many times.  
Under similar exposure, \textsc{Jft} achieves better utility than DPSGD and the redacted models. 

But if the secret detector misses the canary (we purposely code it to mark the canary as public),  the exposure is increased for both ``Redacted (missed)'' and ``\textsc{Jft} (missed)''. But if we add light noise in the first phase (``Redacted+light noise'' in red), even if the canary is missed for 10 times, its exposure is still low. If we continue to privately fine-tune in the second phase (``\textsc{Jft}+light noise'' in pink), we can further improve the utility but still achieve a low exposure value. 
Both DPSGD and CRT also achieve a low exposure value if the canary is missed by the secret detector, but with worse utilities than ``\textsc{Jft}+light noise''. This shows ``\textsc{Jft}+light noise'' can protect missed secrets while achieving better utility. We also performed the canary insertion attack with one canary inserted once, and 10 different canaries inserted once, shown in  Section~\ref{sec:attack in appendix}. 

We also tested the membership inference attack (MIA), but it wasn’t successful (inference accuracy is around 60\% even for public models). Previous studies also observed unsuccessful MIA \cite{shi2021selective, zhao2022provably} and our future work includes developing better MIA for NLP. 

\section{Related Work}

Recent work studied private language models on various model architectures such as RNNs \cite{mcmahan2018learning, ramaswamy2020training} and large language models \cite{anil2021large, li2021large, yu2021differentially}. \citet{li2021large} proposed ghost-clipping to reduce the computational cost in per-sample gradients of DPSGD, and achieved strong private LLMs.  \citet{yu2021differentially} added a small set of private parameters to off-the-shelf LLMs and privately tune them on private data and obtained performant private models. 
Most previous works achieve canonical DP.  \citet{shi2021selective} proposed Selective-DP for applications with sparse sensitive information like NLP, and a privacy mechanism for RNN models. Our work proposes an effective mechanism for LLMs to achieve SDP and study the impact of secret detectors at different levels. 

Our work is also closely related to utilizing public data for  private learning \cite{papernot2018scalable, tramer2020differentially, ghazi2021deep, yu2021differentially}. One working direction assumes access to large unlabeled  public data to train DP models. For example, PATE \cite{papernot2016semi} used unlabeled public data and 
knowledge distillation to build DP models. 
\citet{hoory2021learning} used public medical data to  pre-train  domain-specific private vocabulary and models. Another direction leveraged small public data to guide the private updates in lower-dimension subspaces \cite{zhou2020bypassing, yu2021large}. 
Our work is distinct from previous studies: instead of querying public data from outside, we utilize the public portion of  in-domain data and achieve SDP with better model utility.


\section{Conclusions}
In this paper, we propose \tool, which can achieve Selective-DP for large language models. We also design generalizable secret detectors to provide protection at different levels and study their impacts on the resulting SDP models, and address the problem of missed sensitive tokens via selective manual screening and private training with reduced noise, which is justified by privacy amplification. The results show that the proposed \textsc{Jft} produces SDP models with strong performance while remaining robust to the canary insertion attack. 


\section*{Limitations}
Parameter search in DP learning is challenging as the training process takes a long time and is quite sensitive to different parameters \cite{li2021large}. So the findings in the paper are based on the parameter tuning performed by the authors (Table~\ref{tab:parameter tuning}), and more parameter tuning could potentially lead to better results than the results reported in the paper. 


In our experiments, we simply fine-tuned the models on the in-domain redacted data without adjustment for the redaction. We could potentially utilize more sophisticated methods to train better redacted models to further improve the \textsc{Jft} utility. Besides, for the selective manual screening experiments, we did not adjust the \textit{redacted}-fine-tune step for the low-resource setting where $D'=0.1\%D_0$. Future work includes how to train a better redacted model given limited data. 

When the gap between the SOTA public model and the redacted model is small, the \textit{private}-fine-tuning step cannot further improve the results because of the noisy gradients (e.g., in SST-2), we plan to develop better algorithms to utilize the redacted data and apply denoising methods \cite{welch1995introduction} to close the gap further.


\wyshi{One thing to note is that if the secret detector misses a secret and the secret appears multiple times in the data, then it is likely that the secret detector will miss it multiple times. Therefore, deduplicating the data first is important. We plan to study data deduplication in NLP in the future. }

\section*{Ethical Considerations}
To prevent real-world harm, all the datasets and models used in this paper 
are already public with either public or synthesized  information. So no real personal information will be leaked. 

This work tackles the challenge of privacy protection and can be utilized in various domains or applications to build models that preserve privacy. We will release the code to facilitate privacy-preserving model building. The canary insertion attack is well-known \cite{carlini2019secret,carlini2020extracting} and adjusted specifically for our setting, so it cannot be directly utilized to attack real-world models successfully.   


\section*{Acknowledgements}
We would like to thank Da Yu, Xuechen Li, and Zhiliang Tian for the valuable discussions. We also thank the anonymous area chair and reviewers for their insightful suggestions.

\bibliography{anthology,custom}

\begin{thebibliography}{30}
\expandafter\ifx\csname natexlab\endcsname\relax\def\natexlab#1{#1}\fi

\bibitem[{Abadi et~al.(2016)Abadi, Chu, Goodfellow, McMahan, Mironov, Talwar,
  and Zhang}]{abadi2016deep}
Martin Abadi, Andy Chu, Ian Goodfellow, H~Brendan McMahan, Ilya Mironov, Kunal
  Talwar, and Li~Zhang. 2016.
\newblock Deep learning with differential privacy.
\newblock In \emph{Proceedings of the 2016 ACM SIGSAC conference on computer
  and communications security}, pages 308--318.

\bibitem[{Anil et~al.(2021)Anil, Ghazi, Gupta, Kumar, and
  Manurangsi}]{anil2021large}
Rohan Anil, Badih Ghazi, Vineet Gupta, Ravi Kumar, and Pasin Manurangsi. 2021.
\newblock Large-scale differentially private bert.
\newblock \emph{arXiv preprint arXiv:2108.01624}.

\bibitem[{Balle et~al.(2018)Balle, Barthe, and Gaboardi}]{balle2018privacy}
Borja Balle, Gilles Barthe, and Marco Gaboardi. 2018.
\newblock Privacy amplification by subsampling: Tight analyses via couplings
  and divergences.
\newblock \emph{Advances in Neural Information Processing Systems}, 31.

\bibitem[{Brown et~al.(2022)Brown, Lee, Mireshghallah, Shokri, and
  Tram{\`e}r}]{brown2022does}
Hannah Brown, Katherine Lee, Fatemehsadat Mireshghallah, Reza Shokri, and
  Florian Tram{\`e}r. 2022.
\newblock What does it mean for a language model to preserve privacy?
\newblock \emph{arXiv preprint arXiv:2202.05520}.

\bibitem[{Carlini et~al.(2019)Carlini, Liu, Erlingsson, Kos, and
  Song}]{carlini2019secret}
Nicholas Carlini, Chang Liu, {\'U}lfar Erlingsson, Jernej Kos, and Dawn Song.
  2019.
\newblock The secret sharer: Evaluating and testing unintended memorization in
  neural networks.
\newblock In \emph{28th $\{$USENIX$\}$ Security Symposium ($\{$USENIX$\}$
  Security 19)}, pages 267--284.

\bibitem[{Carlini et~al.(2020)Carlini, Tramer, Wallace, Jagielski,
  Herbert-Voss, Lee, Roberts, Brown, Song, Erlingsson
  et~al.}]{carlini2020extracting}
Nicholas Carlini, Florian Tramer, Eric Wallace, Matthew Jagielski, Ariel
  Herbert-Voss, Katherine Lee, Adam Roberts, Tom Brown, Dawn Song, Ulfar
  Erlingsson, et~al. 2020.
\newblock Extracting training data from large language models.
\newblock \emph{arXiv preprint arXiv:2012.07805}.

\bibitem[{Chen et~al.(2021)Chen, Chen, Yang, Lin, and Yu}]{chen2021action}
Derek Chen, Howard Chen, Yi~Yang, Alex Lin, and Zhou Yu. 2021.
\newblock Action-based conversations dataset: A corpus for building more
  in-depth task-oriented dialogue systems.
\newblock \emph{arXiv preprint arXiv:2104.00783}.

\bibitem[{Doudalis et~al.(2017)Doudalis, Kotsogiannis, Haney, Machanavajjhala,
  and Mehrotra}]{doudalis2017one}
Stelios Doudalis, Ios Kotsogiannis, Samuel Haney, Ashwin Machanavajjhala, and
  Sharad Mehrotra. 2017.
\newblock One-sided differential privacy.
\newblock \emph{Proceedings of the VLDB Endowment}.

\bibitem[{Dwork et~al.(2014)Dwork, Roth et~al.}]{dwork2014algorithmic}
Cynthia Dwork, Aaron Roth, et~al. 2014.
\newblock The algorithmic foundations of differential privacy.
\newblock \emph{Foundations and Trends in Theoretical Computer Science},
  9(3-4):211--407.

\bibitem[{Ghazi et~al.(2021)Ghazi, Golowich, Kumar, Manurangsi, and
  Zhang}]{ghazi2021deep}
Badih Ghazi, Noah Golowich, Ravi Kumar, Pasin Manurangsi, and Chiyuan Zhang.
  2021.
\newblock Deep learning with label differential privacy.
\newblock \emph{Advances in Neural Information Processing Systems}, 34.

\bibitem[{Honnibal and Montani(2017)}]{spacy2}
Matthew Honnibal and Ines Montani. 2017.
\newblock {spaCy 2}: Natural language understanding with {B}loom embeddings,
  convolutional neural networks and incremental parsing.
\newblock To appear.

\bibitem[{Hoory et~al.(2021)Hoory, Feder, Tendler, Erell, Peled-Cohen, Laish,
  Nakhost, Stemmer, Benjamini, Hassidim et~al.}]{hoory2021learning}
Shlomo Hoory, Amir Feder, Avichai Tendler, Sofia Erell, Alon Peled-Cohen, Itay
  Laish, Hootan Nakhost, Uri Stemmer, Ayelet Benjamini, Avinatan Hassidim,
  et~al. 2021.
\newblock Learning and evaluating a differentially private pre-trained language
  model.
\newblock In \emph{Findings of the Association for Computational Linguistics:
  EMNLP 2021}, pages 1178--1189.

\bibitem[{Kingma and Ba(2014)}]{kingma2014adam}
Diederik~P Kingma and Jimmy Ba. 2014.
\newblock Adam: A method for stochastic optimization.
\newblock \emph{arXiv preprint arXiv:1412.6980}.

\bibitem[{Li et~al.(2021)Li, Tramer, Liang, and Hashimoto}]{li2021large}
Xuechen Li, Florian Tramer, Percy Liang, and Tatsunori Hashimoto. 2021.
\newblock Large language models can be strong differentially private learners.
\newblock \emph{arXiv preprint arXiv:2110.05679}.

\bibitem[{Liu et~al.(2019)Liu, Ott, Goyal, Du, Joshi, Chen, Levy, Lewis,
  Zettlemoyer, and Stoyanov}]{liu2019roberta}
Yinhan Liu, Myle Ott, Naman Goyal, Jingfei Du, Mandar Joshi, Danqi Chen, Omer
  Levy, Mike Lewis, Luke Zettlemoyer, and Veselin Stoyanov. 2019.
\newblock Roberta: A robustly optimized bert pretraining approach.
\newblock \emph{arXiv preprint arXiv:1907.11692}.

\bibitem[{McMahan et~al.(2018)McMahan, Ramage, Talwar, and
  Zhang}]{mcmahan2018learning}
H~Brendan McMahan, Daniel Ramage, Kunal Talwar, and Li~Zhang. 2018.
\newblock Learning differentially private recurrent language models.
\newblock In \emph{International Conference on Learning Representations}.

\bibitem[{Merity et~al.(2017)Merity, Xiong, Bradbury, and
  Socher}]{merity2016pointer}
Stephen Merity, Caiming Xiong, James Bradbury, and Richard Socher. 2017.
\newblock Pointer sentinel mixture models.
\newblock \emph{ICLR}.

\bibitem[{Papernot et~al.(2016)Papernot, Abadi, Erlingsson, Goodfellow, and
  Talwar}]{papernot2016semi}
Nicolas Papernot, Mart{\'\i}n Abadi, Ulfar Erlingsson, Ian Goodfellow, and
  Kunal Talwar. 2016.
\newblock Semi-supervised knowledge transfer for deep learning from private
  training data.
\newblock \emph{arXiv preprint arXiv:1610.05755}.

\bibitem[{Papernot et~al.(2018)Papernot, Song, Mironov, Raghunathan, Talwar,
  and Erlingsson}]{papernot2018scalable}
Nicolas Papernot, Shuang Song, Ilya Mironov, Ananth Raghunathan, Kunal Talwar,
  and {\'U}lfar Erlingsson. 2018.
\newblock Scalable private learning with pate.
\newblock \emph{ICLR}.

\bibitem[{Radford et~al.(2019)Radford, Wu, Child, Luan, Amodei, Sutskever
  et~al.}]{radford2019language}
Alec Radford, Jeffrey Wu, Rewon Child, David Luan, Dario Amodei, Ilya
  Sutskever, et~al. 2019.
\newblock Language models are unsupervised multitask learners.
\newblock \emph{OpenAI blog}, 1(8):9.

\bibitem[{Ramaswamy et~al.(2020)Ramaswamy, Thakkar, Mathews, Andrew, McMahan,
  and Beaufays}]{ramaswamy2020training}
Swaroop Ramaswamy, Om~Thakkar, Rajiv Mathews, Galen Andrew, H~Brendan McMahan,
  and Fran{\c{c}}oise Beaufays. 2020.
\newblock Training production language models without memorizing user data.
\newblock \emph{arXiv preprint arXiv:2009.10031}.

\bibitem[{Shi et~al.(2021)Shi, Cui, Li, Jia, and Yu}]{shi2021selective}
Weiyan Shi, Aiqi Cui, Evan Li, Ruoxi Jia, and Zhou Yu. 2021.
\newblock Selective differential privacy for language modeling.
\newblock \emph{arXiv preprint arXiv:2108.12944}.

\bibitem[{Tramer and Boneh(2020)}]{tramer2020differentially}
Florian Tramer and Dan Boneh. 2020.
\newblock Differentially private learning needs better features (or much more
  data).
\newblock \emph{arXiv preprint arXiv:2011.11660}.

\bibitem[{Vaswani et~al.(2017)Vaswani, Shazeer, Parmar, Uszkoreit, Jones,
  Gomez, Kaiser, and Polosukhin}]{vaswani2017attention}
Ashish Vaswani, Noam Shazeer, Niki Parmar, Jakob Uszkoreit, Llion Jones,
  Aidan~N Gomez, {\L}ukasz Kaiser, and Illia Polosukhin. 2017.
\newblock Attention is all you need.
\newblock \emph{Advances in neural information processing systems}, 30.

\bibitem[{Wang et~al.(2018)Wang, Singh, Michael, Hill, Levy, and
  Bowman}]{wang2018glue}
Alex Wang, Amanpreet Singh, Julian Michael, Felix Hill, Omer Levy, and Samuel~R
  Bowman. 2018.
\newblock Glue: A multi-task benchmark and analysis platform for natural
  language understanding.
\newblock \emph{arXiv preprint arXiv:1804.07461}.

\bibitem[{Welch et~al.(1995)Welch, Bishop et~al.}]{welch1995introduction}
Greg Welch, Gary Bishop, et~al. 1995.
\newblock An introduction to the kalman filter.

\bibitem[{Yu et~al.(2021{\natexlab{a}})Yu, Naik, Backurs, Gopi, Inan, Kamath,
  Kulkarni, Lee, Manoel, Wutschitz et~al.}]{yu2021differentially}
Da~Yu, Saurabh Naik, Arturs Backurs, Sivakanth Gopi, Huseyin~A Inan, Gautam
  Kamath, Janardhan Kulkarni, Yin~Tat Lee, Andre Manoel, Lukas Wutschitz,
  et~al. 2021{\natexlab{a}}.
\newblock Differentially private fine-tuning of language models.
\newblock \emph{arXiv preprint arXiv:2110.06500}.

\bibitem[{Yu et~al.(2021{\natexlab{b}})Yu, Zhang, Chen, Yin, and
  Liu}]{yu2021large}
Da~Yu, Huishuai Zhang, Wei Chen, Jian Yin, and Tie-Yan Liu. 2021{\natexlab{b}}.
\newblock Large scale private learning via low-rank reparametrization.
\newblock In \emph{International Conference on Machine Learning}, pages
  12208--12218. PMLR.

\bibitem[{Zhao et~al.(2022)Zhao, Li, and Wang}]{zhao2022provably}
Xuandong Zhao, Lei Li, and Yu-Xiang Wang. 2022.
\newblock Provably confidential language modelling.
\newblock \emph{arXiv preprint arXiv:2205.01863}.

\bibitem[{Zhou et~al.(2020)Zhou, Wu, and Banerjee}]{zhou2020bypassing}
Yingxue Zhou, Zhiwei~Steven Wu, and Arindam Banerjee. 2020.
\newblock Bypassing the ambient dimension: Private sgd with gradient subspace
  identification.
\newblock \emph{arXiv preprint arXiv:2007.03813}.

\end{thebibliography}
\bibliographystyle{acl_natbib}
\newpage
\appendix

\section{Appendix}
\label{sec:appendix}
\subsection{Proofs}
\label{sec:proof}

\begin{customthm}{1}[restated]
Given that 1) in the first phase, the data used for fine-tuning does not contain sensitive tokens and a public optimizer is used, and 2) in the second phase, the private optimizer achieves $(\epsilon,\delta)$-DP, \tool achieves $(\epsilon, \delta)$-SDP.
\end{customthm}

\begin{proof}
Since the first phase does not incur any privacy loss on the sensitive tokens, the first phase achieves $(0,0)$-SDP.

DP implies SDP. In other words, if an algorithm achieves $(\epsilon,\delta)$-DP, then it also satisfies $(\epsilon,\delta)$-SDP. Hence, the second phase achieves $(\epsilon,\delta)$-SDP. By the composition of SDP, \tool achieves $(\epsilon,\delta)$-SDP.

\wyshi{Note that the first phase achieves (0,0)-SDP but cannot achieve (0,0)-DP. DP aims to protect the entire token sequence, whether it is considered sensitive or non-sensitive by the policy function. Because the first phase does not noise the non-sensitive tokens at all, it cannot ensure DP.
}
\end{proof}

\subsection{Implementation Details}
\label{sec:implementation detail}
\noindent\textbf{Notes on special tokens.} When fine-tuning LLMs, it is a common practice to add new special tokens to fit the need of the downstream tasks. For example, in dialogue tasks, we often have prompts like ``SYS:'' and ``USR:'' to indicate the speaker in the data. This step doesn't affect the public models that much (20.48 without special tokens vs 20.44 with special tokens), 
but as it does change the model structure (additional embeddings) and the model initialization, 
we notice that in our experiments, DPSGD is sensitive to the addition of special tokens (because the model initialization is changed): after reasonable amounts of parameter tuning (see Table~\ref{tab:parameter tuning}), 
DPSGD initialized with the original GPT achieves 27.05 in PPL, 
while DPSGD with added special tokens achieves 30.32 in PPL on Wikitext-2. The gap could potentially be reduced with more parameter tuning, but we just want to mention that in practice, it may not be easy to find the best parameters. In our experiments, for WikiText-2, we add \textit{$<$mask$>$} as the special token; for ABCD, as it is a dialogue task, we add \textit{$<$mask$>$, ``ACT:'', ``SYS:''}, and \textit{``USR:''}.  
Since all the \textsc{Jft} models have added special tokens, we report two DPSGD results, one without special tokens and one with special tokens,  for a fair comparison in terms of model structure. 

Also, the secret detectors replace the sensitive information with artificial special tokens such as ``$<$SSN$>$'' and ``$<$NAME$>$''.  
But these tokens don't appear in the validation or test set and thus inserting them will skew the training data distribution and lead to inferior results, 
especially when the sensitive token portion is high.  
In our experiments, we mask the detected sensitive information with the same ``$<$mask$>$'' token  and ignore this special token in the loss calculation.
In this way, for models with an existing ``$<$mask$>$'' token (like Roberta), we can utilize the existing embedding; for models without ``$<$mask$>$'',  the model only needs to learn  one additional special embedding. 
This improves the validation perplexity from $64.82$ to $37.90$ for the redacted GPT2 model with the low contextual secret detector. 

\wyshi{We could potentially apply the same secret detector on the validation and test set to mitigate special token issues. However, this causes two concerns: 1) if the secrete detector redacts 45\% of tokens (e.g. the high contextual one redacts all the verbs, etc), then the performance on validation/test is not informative at all, and cannot be compared to the public baseline; 2) in the past privacy literature \cite{papernot2018scalable, ghazi2021deep}, the conventional problem setup considers validation/test sets as public and focuses only on the training privacy. We inherit the same treatment to be comparable with prior literature. But in privacy-related NLP problems, how to treat the validation/test sets remains an open question as they can contain private information.}

\wyshi{Our experiments find that adding many special tokens impacts the results. In the future, we plan to study how to  treat special tokens better in privacy-preserving LMs. 
}
\begin{table}[!htbp]
\centering
\begin{tabular}{l|>{\centering\arraybackslash}p{0.5\linewidth}}
\toprule
\textbf{Parameters}           & \textbf{Range}                                    \\
\midrule
$\epsilon$           & 3                                        \\
Clipping norm $C$      & 0.1                                      \\
Batch size           & \{256, 512, 1024\}                       \\
Learning rate        & \{5, 10, 50\} $\cdot 10^{-5}$ \\
Epochs $E$             & \{10, 100, 200, 600\}                    \\
Noise scale $\sigma$ & Pre-calculated so that $\epsilon$ is spent when training ends                          \\
\bottomrule
\end{tabular}
\caption{Hyper-parameter tuning range.}
\label{tab:parameter tuning}
\end{table}

\begin{table*}[!htbp]
    \centering
    \resizebox{\textwidth}{!}{
    \begin{tabular}{l|l|cc|cc|cc|cc||cc|cc}
    \midrule
   & &\multicolumn{2}{c|}{\textbf{MNLI}} & \multicolumn{2}{c|}{\textbf{\textsc{QQP}}}&
   \multicolumn{2}{c|}{\textbf{\textsc{QNLI}}}&
   \multicolumn{2}{c||}{\textbf{\textsc{SST-2}}}&
   \multicolumn{2}{c|}{\textbf{\textsc{WikiText-2}}}&
   \multicolumn{2}{c}{\textbf{\textsc{ABCD}}}
   \\
\midrule
 \textbf{Model}
&\textbf{Detector}& \textbf{Acc$\uparrow$} 
&  \textbf{Privacy} &   
 \textbf{Acc$\uparrow$} 
&  \textbf{Privacy} &
 \textbf{Acc$\uparrow$} 
&  \textbf{Privacy} &
 \textbf{Acc$\uparrow$} 
&  \textbf{Privacy} &
 \textbf{PPL$\downarrow$} 
&  \textbf{Privacy} &
 \textbf{PPL$\downarrow$} 
&  \textbf{Privacy} 
 \\ 
 \midrule

CRT & low ent&  81.45 & (2.21, $\delta_c$)-Conf &   84.10 & (2.47, $\delta_c$)-Conf &   83.36 & (2.30, $\delta_c$)-Conf &  87.39 & (2.22, $\delta_c$)-Conf &  28.20 & (2.19, $\delta_c$)-Conf &  9.09 & (2.73, $\delta_c$)-Conf\\

\textsc{Jft} & low ent &  \textbf{85.74} & (0.92, $\delta_s$)-SDP  &  \textbf{88.19} & (2.58, $\delta_s$)-SDP &  \textbf{89.57} & (2.37, $\delta_s$)-SDP &  \textbf{92.09} & (2.06, $\delta_s$)-SDP &  \textbf{21.86} & (2.58, $\delta_s$)-SDP &  \textbf{6.09} & (2.71, $\delta_s$)-SDP
\\

\midrule

CRT & high ent &  81.24 & (2.63, $\delta_c$)-Conf &  83.64 & (2.13, $\delta_c$)-Conf &   82.78 & (2.56, $\delta_c$)-Conf &  87.27 & (2.22, $\delta_c$)-Conf &  28.47 & (1.38, $\delta_c$)-Conf &  9.20 & (2.71, $\delta_c$)-Conf
\\

\textsc{Jft} & high ent &  \textbf{85.61} & (0.99, $\delta_s$)-SDP &  \textbf{88.05} & (2.58, $\delta_s$)-SDP &  \textbf{89.35} & (2.37, $\delta_s$)-SDP &  \textbf{92.20} & (2.12, $\delta_s$)-SDP&  \textbf{22.55} & (2.58, $\delta_s$)-SDP &  \textbf{6.25} & (2.71, $\delta_s$)-SDP
\\
\midrule
CRT & low ctx &  78.85 & (2.46, $\delta_c$)-Conf &  79.17 & (2.50, $\delta_c$)-Conf &  81.15 & (2.24, $\delta_c$)-Conf &  85.67 & (2.22, $\delta_c$)-Conf &  28.87 & (0.69, $\delta_c$)-Conf &  12.70 & (0.34, $\delta_c$)-Conf
\\

\textsc{Jft} & low ctx &  \textbf{85.02} & (1.23, $\delta_s$)-SDP &  \textbf{87.00} & (2.41, $\delta_s$)-SDP &  \textbf{87.99} & (2.52, $\delta_s$)-SDP &  \textbf{92.43} & (2.17, $\delta_s$)-SDP&  \textbf{25.62} & (2.58, $\delta_s$)-SDP &  \textbf{8.80} & (2.71, $\delta_s$)-SDP
\\
\midrule
\multicolumn{14}{l}{\textbf{Stress-test}}        \\        
\midrule
CRT & high ctx &  74.61 & (2.68, $\delta_c$)-Conf &  77.57 & (2.64, $\delta_c$)-Conf &  79.41 & (2.30, $\delta_c$)-Conf &  86.01 & (2.33, $\delta_c$)-Conf &  29.13 & (0.47, $\delta_c$)-Conf &  13.11 & (0.35, $\delta_c$)-Conf\\

\textsc{Jft} & high ctx &  \textbf{84.11} & (1.18, $\delta_s$)-SDP &  \textbf{86.42} & (2.67, $\delta_s$)-SDP &  \textbf{87.06} & (2.41, $\delta_s$)-SDP &  \textbf{91.17} & (2.17, $\delta_s$)-SDP&   \textbf{27.19} & (1.96, $\delta_s$)-SDP &  \textbf{12.93} & (2.71, $\delta_s$)-SDP\\

\bottomrule

\end{tabular}
}
    \caption{Comparison between \textsc{Jft} and CRT. \textsc{Jft} achieves $(\epsilon_s, \delta_s)$-\emph{Selective-DP}, while CRT achieves $(\epsilon_c, \delta_c)$-\emph{Confidentiality}, so the $\epsilon$ are \emph{not} directly comparable. For QNLI and SST-2, $\delta_s=\delta_c\approx$1e-5; for MNLI and QQP, $\delta_s=\delta_c\approx$1e-6. For generation task, $\delta_s=\delta_c\approx$1e-6.  We add the same amount of noise to \textsc{Jft} and CRT (noise calculated based on $\epsilon$=3), and report the best model validation utility.  
    ``CRT'' results are based on our implementation of \citet{zhao2022provably}. 
    }
    \label{tab:CRT vs SDP}
\end{table*}

\begin{table}[!htbp]
\centering
\begin{tabular}{c|c|c}
\toprule
\textbf{Batch size}           & \textbf{Epoch} & \textbf{PPL}                                    \\
\midrule
256 & 200 & 27.04\\
512 & 200 & 27.05\\
1024 & 200 & 27.18\\
1024 & 600 & 27.01\\
1024 & 10 & 28.84\\
\bottomrule
\end{tabular}
\caption{Model utility with different hyper-parameters.}
\label{tab:ppl diff hyper}
\vspace{-1em}
\end{table}

\noindent\textbf{Hyper-parameter tuning.} Hyper-parameter tuning remains a challenging problem in DP learning as the training takes a long time, and the model can be sensitive to the hyper-parameters. Guided by \citet{li2021large}, we tune the learning rate, the number of training epochs, and the batch size on the validation set and report the best results. Please refer to Table~\ref{tab:parameter tuning} for the parameter range.

\wyshi{In our experiments, we tuned the batch size, learning rate, and epoch number for the best models in different settings. In practice, we found increasing the epoch number and batch size helps, but these two factors can interact with each other, the gain could be small after the batch size and epoch number are large enough. Table~\ref{tab:ppl diff hyper} shows the convergent model utility of DPSGD on WikiText-2 with different hyper-parameters. In our experiments, we pick the best set of parameters that balances the computational cost and the utility.  
}


\subsection{Comparison with CRT}
Table~\ref{tab:CRT vs SDP} shows the comparison between \textsc{Jft} and CRT on the model utility and privacy guarantee.  Note that \textsc{Jft} achieves $(\epsilon_s, \delta_s)$-SDP, and CRT realizes $(\epsilon_c, \delta_c)$-Confidentiality, because the underlying privacy notion is different,  we cannot directly compare the $\epsilon$. We pre-calculate the noises given $\epsilon=3$, add the same amount of noises to both CRT and \textsc{Jft}, and report the best valid utility. With the same amount of noise added, \textsc{Jft} achieves better model utilities than CRT for all the tasks across different secret detectors.

\subsection{Low-Resource Results}
\label{sec:low-resource}

Privacy protection is  important in oftentimes low-resource domains such as health care. 
We simulate the low-resource setting where we have both limited redacted and private data, i.e., ($D'$=0.1\% $D_0$, $D$=0.1\% $D_0$) and the results are in Table~\ref{tab:complete-low-resource}.  

The redacted model always performs better than DPSGD, suggesting that for low-resource settings, we can simply redact the data to train the model instead of employing differential privacy. Also, for the  QNLI task, \textsc{Jft} shows promising results.  With 0.1\% training data (100 records), the redacted model improve the accuracy from random guess to 66.5\%. \textsc{Jft} can even further improve the accuracy to 67.49\%. But the baseline DPSGD fails to improve the model at all (accuracy=50.54\%). We plan to study how to better fine-tune the redacted model privately with limited data.  

\begin{table}[htbp!]
\centering
\begin{adjustbox}{width=.95\columnwidth}
\begin{tabular}{l|cccccc}
\toprule

\textbf{Manual Screening}& \multicolumn{6}{c}{\textbf{$D'$=0.1\%$D_0$, $D$=0.1\%$D_0$}}                                           \\
\midrule

\multirow{2}{*}{\textbf{Task}}  
& \textbf{MNLI} & \textbf{QQP}   & \textbf{QNLI} & \textbf{SST-2} & \textbf{WikiText-2} & \textbf{ABCD} \\


& Acc $\uparrow$ & Acc $\uparrow$ & Acc $\uparrow$ & Acc $\uparrow$ & PPL $\downarrow$ & PPL $\downarrow$ \\

 \midrule

\textbf{$D'$} 
& 300 & 300   & 100 & 100 & 10 & 10 \\

 \midrule

\textbf{DPSGD} 
& 32.86 &63.18  & 50.54   &56.77 & 30.08  & 13.66 
\\
\midrule
\textbf{Redacted} 
& \textbf{52.52} & \textbf{75.25} & 66.48 & \textbf{88.88} & \textbf{28.06} & \textbf{9.36} 
\\
\textbf{\textsc{Jft}+manual screening}  
& 50.61 & 75.18 & \textbf{67.49} & 88.53  & 28.15    &9.40
\\
\bottomrule
\end{tabular}
\end{adjustbox}
\caption{Manual screening results when both $D$ and $D'$ are limited in size. $D'$ is the redacted data, $D$ is the private data. 
}
\label{tab:complete-low-resource}
\vspace{-1em}
\end{table}

\subsection{Canary Insertion Attack}
\label{sec:attack in appendix}

Figure~\ref{fig:canary_1} shows the canary insertion attack result when the canary is inserted only once. We see that the exposure is low for all the models (<3, so not extractable) including the public ``No-DP'' without any protections. This agrees with Figure~4 in \citet{carlini2019secret} that when the canary is inserted for very limited times, its exposure is low. 
\begin{figure}[htbp!]
    \centering
    \includegraphics[scale=0.32]{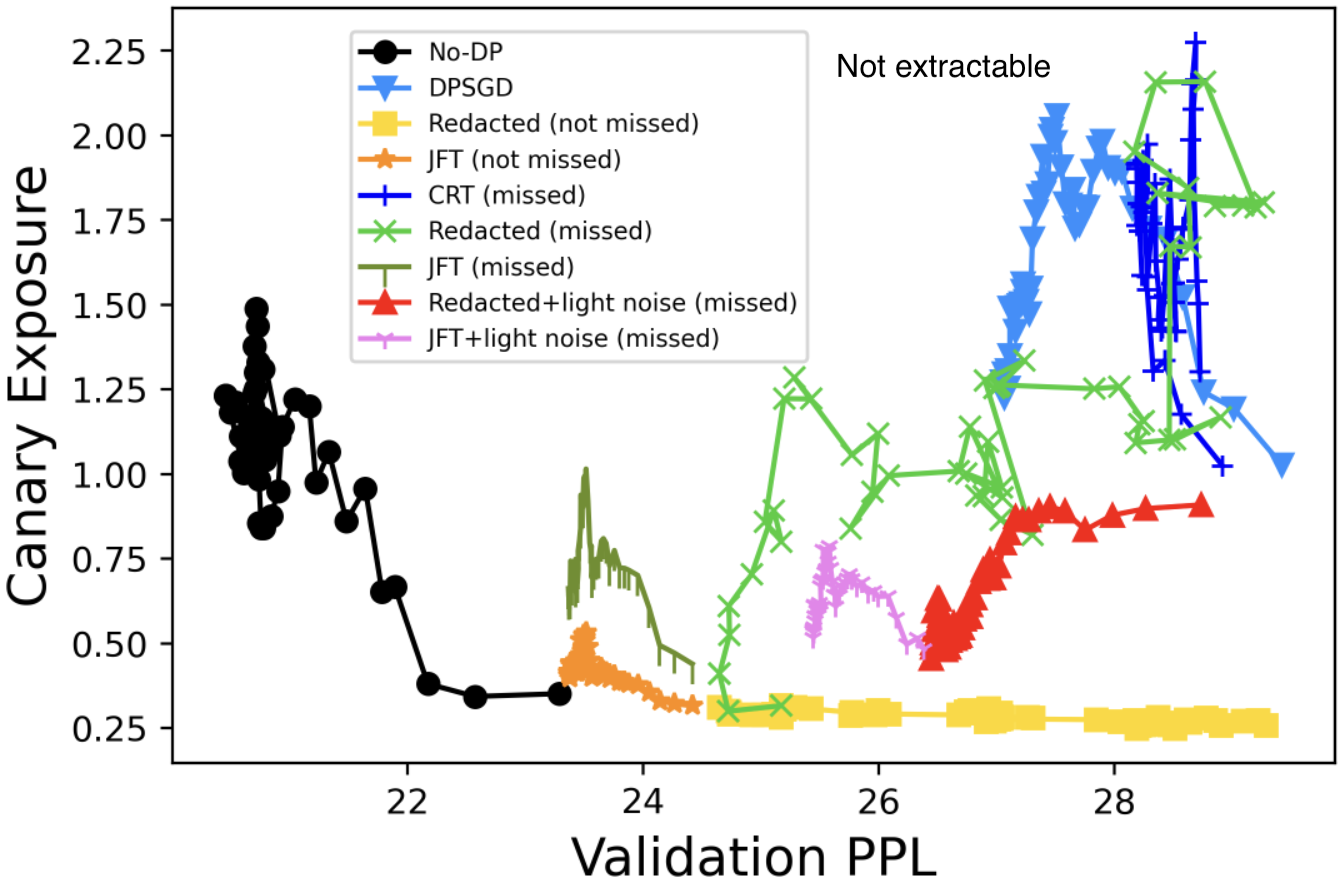}
    \caption{Exposure for different models when the canary is inserted only once. The exposures are all small (<3) even for public models. }
    \label{fig:canary_1}
\end{figure}

\begin{figure}[htbp!]
    \centering
    \includegraphics[scale=0.32]{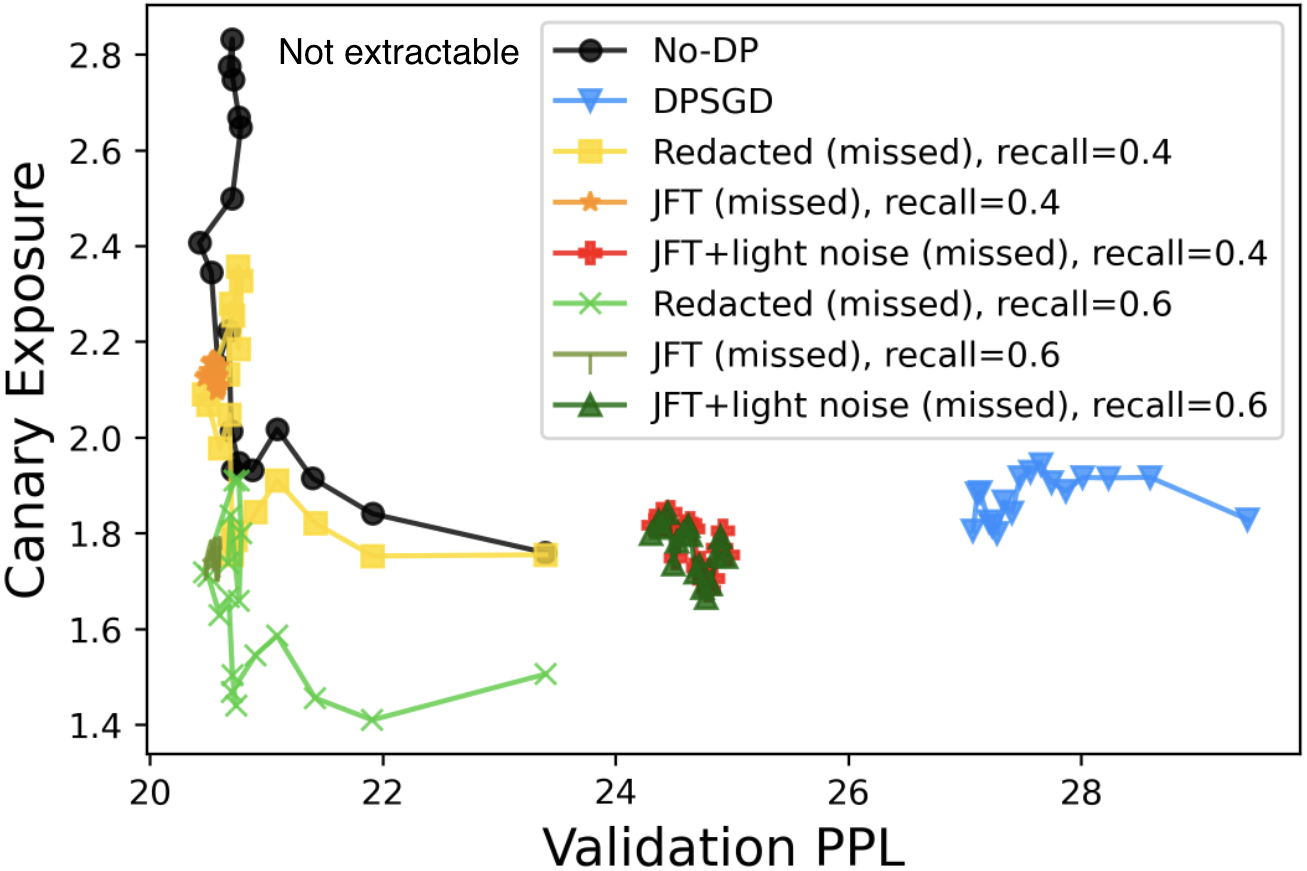}
    \caption{Exposure for different models when we insert ten different canaries.}
    \label{fig:canary_10_diff}
    \vspace{-1em}
\end{figure}

\wyshi{Figure~\ref{fig:canary_10_diff} shows the canary insertion attack results when we insert ten different canaries into the training data. The exposure is the average exposure of the ten canaries. In this experiment, we treat the inserted canaries as the only secrets, so the ``Redacted'' and ``\tool'' model utilities are close to the black ``No-DP'' model, and we can better compare with the ``No-DP'' model. We also artificially vary the recall of the secret detector to see the effect. ``Recall=0.4'' means that the detector can only detect four of the  ten canaries. Because each canary only appears once in the dataset, the exposure is low, similar to the ones in Figure~\ref{fig:canary_1}. But if the recall is higher (0.6), the exposure will still be lower. ``\tool+light noise'' achieves low exposure, similar to the baseline DPSGD that protects all canaries, but with much higher utility over DPSGD. }

\wyshi{These experiments may suggest that for large NLP models, if the sensitive tokens only appear for very limited times, they may not be extracted using the canary insertion attack. }

\end{document}